\documentclass{article}

\usepackage[preprint,nonatbib]{neurips_2026}
\usepackage[numbers,sort&compress]{natbib}

\usepackage[utf8]{inputenc}
\usepackage[T1]{fontenc}
\usepackage{lmodern}
\usepackage{microtype}
\usepackage{amsmath, amssymb}
\usepackage{graphicx}
\usepackage{booktabs}
\usepackage{array}
\usepackage{xcolor}
\usepackage{enumitem}
\usepackage{hyperref}
\usepackage{url}
\usepackage{nicefrac}
\usepackage{caption}
\usepackage{float}

\newcommand{\Neff}{N_{\mathrm{eff}}}
\newcommand{\aZ}{\alpha_{\mathrm{Z}}}
\newcommand{\mmax}{m_{\mathrm{max}}}

\title{Predictable Confabulations: Factual Recall by LLMs Scales with Model Size and Topic Frequency}

\author{%
  Matthew L. Smith\thanks{Corresponding author: \texttt{msmith@idrc.ca}} \\
  International Development Research Centre\\
  Canada\\
  \And
  Jonathan P. Shock \\
  University of Cape Town\\
  South Africa\\
  \And
  Samuel T. Segun \\
  Global Center on AI Governance\\
  Canada\\
  \texttt{hello@samuelsegun.com}\\
  \And
  Iyiola E. Olatunji \\
  SnT, University of Luxembourg\\
  Luxembourg\\
  \And
  Tegawend\'{e} F. Bissyand\'{e} \\
  SnT, University of Luxembourg\\
  CITADEL AI Centre of Excellence, Burkina Faso\\
  \texttt{tegawende.bissyande@uni.lu}\\
}

\begin{document}

\maketitle

\begin{abstract}
While scaling laws govern aggregate large language model performance, no scaling law has linked factual recall to both model size and training-data composition. We evaluated 38 models on over 8{,}900 scholarly references evaluated by an automated reference verification system. Recall quality follows a sigmoid in the log-linear combination of model parameter count and topic representation in training data. These two variables alone explain 60\% of the variance across 16 dense models from four families, rising to 74--94\% within individual families. The form matches a superposition-inspired account in which recall is gated by a signal-to-noise ratio: signal strength scales with concept frequency and the noise floor with model capacity.
\end{abstract}


\section{Introduction}

Factual recall errors, or confabulations, remain one of the most consequential failure modes of large language models (LLMs)\cite{kalai2025}. Prior work has linked these failures to both model scale and the distribution of knowledge in training data, but no single framework combines both factors into a predictive scaling law for factual recall quality.

We measured recall quality across 38 LLMs and 24 topics spanning five orders of magnitude in training-data representation. Using academic references as the factual recall being measured, we find that recall quality scales with the log of a signal-to-noise ratio (SNR), where signal strength is set by how frequently a concept appears in training data and the noise floor is set by model capacity. Under a superposition-inspired account, both contributions enter the SNR multiplicatively, yielding a sigmoid in their log-sum:
\begin{equation}
  \text{quality} = \sigma\!\left(\alpha\,\log_{10} P + \beta\,\log_{10} S + \gamma\right)
  \label{eq:model}
\end{equation}
where $P$ is model parameter count, $S$ is a proxy for concept frequency in training data, and $\sigma$ is the logistic function. Fitted to 16 dense models across 24 topics ($N = 384$ model--topic observations from 3{,}661 evaluated references; $R^2 = 0.599$), the relationship holds across independent model families. Across all 38 models and 8{,}913 evaluated scholarly references, topic representation frequency is positively associated with recall quality independently of architecture or parameter count.

The remainder of this paper is organised as follows. Section~\ref{sec:related} reviews the relevant literature. Section~\ref{sec:theory} develops the theoretical framework that motivates the sigmoid functional form. Section~\ref{sec:methods} describes the experimental design, verification pipeline, and statistical methods. Section~\ref{sec:results} presents the empirical findings, and Section~\ref{sec:discussion} interprets them. Section~\ref{sec:conclusion} summarises our contributions and outlines directions for future work.


\section{Related Work}
\label{sec:related}

\paragraph{Scaling laws for language models.}
Kaplan et al.\cite{kaplan2020} established that aggregate language model performance improves as a power law in model size, dataset size, and compute. Hoffmann et al.\cite{hoffmann2022} refined these relationships to show that model and dataset size should be scaled in tandem for compute-optimal training. These aggregate scaling laws describe loss on held-out text but do not decompose performance into specific capabilities such as factual recall.

\paragraph{Knowledge capacity and superposition.}
Allen-Zhu and Li\cite{allenzhu2024} showed that knowledge capacity scales at roughly two bits per parameter, providing a quantitative link between model size and the number of facts that can be stored. Mechanistically, Elhage et al.\cite{elhage2022} introduced the concept of superposition---the encoding of more features than a network has dimensions---and demonstrated it in toy models. Bricken et al.\cite{bricken2023} and Templeton et al.\cite{templeton2024} confirmed that superposition operates in production language models via sparse autoencoder analysis. Bereska et al.\cite{bereska2024superposition} formalised superposition as lossy compression. Liu et al.\cite{liu2025superposition} provided empirical evidence across four open-source LLM families that models operate in a strong-superposition regime where representation vectors form near-equiangular tight frames with squared overlaps scaling as $1/d$ (where $d$ is model width).

\paragraph{Knowledge frequency and the long tail.}
A complementary line of work has established that rarer concepts are disproportionately susceptible to recall failure. Kandpal et al.\cite{kandpal2023longtail} showed that LLMs struggle to learn long-tail knowledge, with question-answering accuracy strongly correlated with the number of training documents containing the relevant entity. Mallen et al.\cite{mallen2023trust} found that parametric memory is reliable only for popular entities, while retrieval augmentation is needed for the long tail. Sun et al.\cite{sun2024headtotail} confirmed this head-to-tail degradation across a range of LLMs, and Lu et al.\cite{lu2024scaling} derived scaling laws for fact memorisation as a function of model size and training epochs. Yin et al.\cite{yin2025overshadowing} formalised the phenomenon as a ``law of knowledge overshadowing,'' in which high-frequency knowledge suppresses retrieval of related low-frequency facts. Yuan et al.\cite{yuan2024} developed holistic benchmarks for factual knowledge recall. Moayeri et al.\cite{moayeri2024worldbench} demonstrated systematic geographic biases in LLM factual recall, consistent with frequency-driven disparities.

\paragraph{Memorisation and training data.}
Carlini et al.\cite{carlini2023quantifying} quantified memorisation across neural language models, showing that larger models memorise more training data and that memorisation increases with the number of times a sequence appears in training. Hernandez et al.\cite{hernandez2022scaling} established scaling laws for learning from repeated data, finding that data repetition and model size interact in predictable ways. Xiong et al.\cite{xiong2025sok} provide a comprehensive survey of memorisation mechanisms, measurement, and mitigation in LLMs.

\paragraph{This work.}
While these strands have established model size and knowledge frequency as independent drivers of factual recall, no prior work has combined them into a single predictive functional form. We provide this unification by showing that a sigmoid in the log-linear combination of both factors---motivated by a superposition-inspired SNR account---fits the data across four independent model families.


\section{Theoretical Framework}
\label{sec:theory}

The functional form fitted in Section~\ref{sec:results} is motivated by two ingredients: the geometry of superposition in neural representations, and the statistics of concept frequencies in natural-language training corpora. We present this as a superposition-inspired latent signal-to-noise model rather than a derivation: each step involves an explicit assumption that we flag below, and the data are consistent with the resulting sigmoid form rather than uniquely selecting it.

\subsection{Setup: superposition, interference, and the recall threshold}

We use the following notation throughout. On the \emph{model side}, $P$ is the total parameter count and $\Neff$ the effective number of dimensions available for factual encoding, with $\Neff \ll P$ and $\Neff \propto P^{\gamma_e}$ for some $\gamma_e \leq 1$. $M$ is the number of factual concepts encountered in training.

On the \emph{data side}, $f_i$ is the training-corpus frequency of concept~$i$. The encoded signal strength of concept~$i$ is $s_i(f_i)$, a monotonically increasing function of $f_i$ for which we adopt the simplest scale-free form $s(f) = f^{\beta_s}$. (Note that $\beta_s$ and $\gamma_e$ are theoretical exponents distinct from the fitted parameters $\beta$ and $\gamma$.) $S$ denotes the OpenAlex scholarly work count for a topic, serving as an empirical proxy for training-data representation at the topic level, with individual concept frequencies scaling as $f \propto S^\delta$ for some $\delta \leq 1$. We note that $S$ also reflects the number of distinct facts associated with a topic (its inventory size), so the $f \propto S^\delta$ assumption with $\delta < 1$ partially absorbs the breadth-versus-reinforcement ambiguity but does not formally separate signal strength from concept count; the framework should accordingly be read as a coarse-grained model in which $S$ acts as a single composite proxy.

When $M > \Neff$, the network must encode concepts in overlapping directions, a phenomenon termed superposition\cite{elhage2022} and confirmed by sparse-autoencoder studies\cite{bricken2023, templeton2024} as well as work formalising it as lossy compression\cite{bereska2024superposition}. Let $\mathbf{v}_i \in \mathbb{R}^{\Neff}$ denote the representation vector for concept~$i$. Retrieving concept $i$ produces a readout
\begin{equation}
  r_i = \langle \mathbf{v}_i, \mathbf{v}_i \rangle
      + \sum_{j \neq i} \langle \mathbf{v}_i, \mathbf{v}_j \rangle,
  \label{eq:retrieval}
\end{equation}
where $r_i$ is the retrieval signal, the first term is the true signal, and the second term is interference from the remaining concepts. For near-random vectors in $\mathbb{R}^{\Neff}$, the expected squared inner product is $\sim 1/\Neff$, so the typical pairwise overlap scales as $\sim 1/\sqrt{\Neff}$. We use $\varepsilon(\Neff) \sim 1/\sqrt{\Neff}$ as a coarse summary of the interference floor; the aggregate interference at retrieval depends additionally on the effective number of active competitors and their relative signal strengths, so this expression should be read as a typical-case scaling rather than a tight bound. Liu et al.\cite{liu2025superposition} provide empirical support across four LLM families, showing that open-source LLMs operate in a ``strong superposition'' regime where representation vectors form near-equiangular tight frames with squared overlaps scaling as $1/d$ (where $d$ is model width).

A concept is recalled when its signal exceeds this interference floor,
\begin{equation}
  s_i(f_i) > \varepsilon(\Neff) \sim \frac{1}{\sqrt{\Neff}},
  \label{eq:threshold}
\end{equation}
so the signal-to-noise ratio $\mathrm{SNR}_i = s_i(f_i) / \varepsilon(\Neff)$ controls recall. Increasing $\Neff$ (via more parameters) lowers the floor; increasing $f_i$ (via more training data on a topic) raises the signal.

\subsection{Scaling law for the recall fraction}

Natural-language concept frequencies are well approximated by a power law (Zipf's law): if concepts are ranked by frequency, the $k$-th concept has frequency $f_k \propto k^{-\aZ}$, where $\aZ > 1$ is the Zipf exponent. With this assumption and the power-law signal mapping $s(f) = f^{\beta_s}$, the threshold condition (Eq.~\ref{eq:threshold}) becomes $k^{-\aZ\beta_s} > \Neff^{-1/2}$, which is satisfied for all ranks $k < k^*$ where $k^* \propto \Neff^{1/(2\aZ\beta_s)}$. With $\Neff \propto P^{\gamma_e}$, the recall fraction $Q = k^*/M$ scales as
\begin{equation}
  Q \propto P^{\,\gamma_e/(2\aZ\beta_s)}.
  \label{eq:scaling}
\end{equation}
A worked estimate from Liu et al.\cite{liu2025superposition} is illustrative: they measure $P \sim d^{2.52}$ across LLM families, giving $\gamma_e \approx 0.40$ if $\Neff$ is identified with model width~$d$. (Total encoding capacity likely also involves depth, so this is a lower bound on $\gamma_e$.)

For the topic axis, we incorporate the empirical proxy $S$ by writing the frequency of the $k$-th concept within a topic as $f_k = c_0 \cdot S^\delta \cdot k^{-\aZ}$, where $\delta < 1$ because additional papers on a topic typically introduce new facts rather than reinforce existing ones. Substituting into the threshold condition:
\[
  (c_0 \cdot S^\delta \cdot k^{-\aZ})^{\beta_s} > \Neff^{-1/2}
  \quad\Longrightarrow\quad
  k^{\aZ\beta_s} < c_0^{\beta_s} \cdot S^{\delta\beta_s} \cdot \Neff^{1/2}.
\]
Taking the $S$ dependence at fixed $P$ (and hence fixed $\Neff$), we get $k^* \propto S^{\delta\beta_s/(\aZ\beta_s)} = S^{\delta/\aZ}$. The exponent $\beta_s$ cancels in this $S$ dependence (though it remains in the $P$ dependence of Eq.~\ref{eq:scaling}), so
\begin{equation}
  Q \propto S^{\,\delta/\aZ}.
  \label{eq:content}
\end{equation}
Combining the two axes,
\begin{equation}
  Q \;=\; C \cdot P^{\,\gamma_e/(2\aZ\beta_s)} \cdot S^{\,\delta/\aZ}.
  \label{eq:Q-power-law}
\end{equation}
The three power-law assumptions ($s \propto f^{\beta_s}$, $\Neff \propto P^{\gamma_e}$, $f \propto S^\delta$) cannot be independently verified from our data: only the compound exponents are observable, absorbed into the fitted $\alpha$ and $\beta$ of Equation~\ref{eq:model}. All three must be power laws for the log-linear form to hold, and the high $R^2$ of within-family fits constrains how far they can deviate. At very small model sizes the $\Neff \propto P^{\gamma_e}$ assumption likely breaks down: the floor regime, in which models produce template fabrications rather than noisy recall, is consistent with $\Neff$ falling effectively to zero below a family-specific threshold.

\subsection{Mapping to continuous quality}

The recall fraction $Q = k^*/M$ is bounded in $[0,1]$ by definition, but Equation~\ref{eq:Q-power-law} can exceed~1 for sufficiently large $P$ and $S$, indicating a regime where capacity exceeds the concept inventory and quality saturates. Our experiment measures a continuous quality score on $[0,1]$ via free recall: each model produces 10 references per topic, and we score field-by-field accuracy. The measured quantity is therefore an order-statistic over the model's most-confident outputs, not the population recall fraction $Q$ itself; if the model can rank concepts by internal signal strength, easy/canonical references will be over-represented and the empirical mapping inherits any selection-induced bias. We use a logistic link as a tractable, bounded mapping between the latent $\log Q$ and observed quality, motivated by but not derived from the superposition account:
\begin{equation}
  \text{quality} \;=\; \sigma\!\bigl(a \cdot \log_{10} Q \;+\; b\bigr),
  \label{eq:sigmoid}
\end{equation}
where $\sigma(x) = 1/(1+e^{-x})$ and $a, b$ are empirical constants.

This produces three regimes. For $\log Q \ll 0$ (the \emph{floor}), $\sigma(x) \approx e^x$, so quality $\propto Q^{\,a/\ln 10}$---a power law approaching zero. Near the inflection point (the \emph{ramp}), $\sigma$ is approximately linear, yielding the tangent-line form
\begin{equation}
  \text{quality} \approx m \cdot \log_{10}(P) + n \cdot \log_{10}(S) + c,
  \label{eq:quality-model}
\end{equation}
which is the small-curvature limit of Equation~\ref{eq:model}. The slope of the logistic at the inflection point is $\sigma'(0) = 1/4$, so $m = \alpha/4$, $n = \beta/4$, and $c = 1/2 + \gamma/4$. For $\log Q \gg 0$ (the \emph{ceiling}), $\sigma(x) \approx 1 - e^{-x}$, so the distortion $1 - \text{quality} \propto P^{-m'} S^{-n'}$, where $m', n'$ inherit the factor $a/\ln 10$ from the sigmoid argument. This recalls the rate--distortion framework of Shannon\cite{shannon1959}, in which distortion decays exponentially with rate ($D \propto 2^{-2R}$ for a Gaussian source); the ceiling regime of our logistic mapping takes the same form with an effective exponent set by the underlying scaling parameters.

The fitted parameters $\alpha$ and $\beta$ in Equation~\ref{eq:model} therefore absorb both the theoretical exponents ($\gamma_e, \aZ, \beta_s, \delta$) and the linearisation factor $a$. The sigmoid derivative $\sigma'(z) = q(1-q)$, with $q$ denoting quality, predicts an inverted-U in the effective content sensitivity $n_{\text{eff}} \equiv \partial\,\text{quality} / \partial \log_{10} S$: it peaks at $q \approx 0.5$ and declines toward both extremes.

\subsection{Empirical constraints on the Zipf exponent and the reference slope}

From the OpenAlex concept hierarchy (65,026 concepts), OLS regression yields $\aZ = 1.23$ with bootstrap 95\% CI $[1.18, 1.30]$, confirmed by maximum-likelihood estimation\cite{clauset2009} at $\aZ = 1.24$. Independently, Michaud et al.\cite{michaud2023quantization} measured $\aZ \approx 1.24$ for the rank--frequency distribution of learned ``quanta'' in a Pythia model---a strikingly close agreement, given that one estimate is derived from external scholarly concept frequencies and the other from internal model behaviour. A rolling-window analysis reveals that $\aZ$ varies from ${\sim}1.0$ at broad fields to ${\sim}3$--$4$ at niche topics.

The threshold derivation (Eq.~\ref{eq:scaling}) yields a slope $\gamma_e/(2\aZ\beta_s)$ for $\log Q$ vs $\log P$. Setting the canonical reference values $\gamma_e = \beta_s = 1$ gives a benchmark slope $\mmax = 1/(2\aZ) \approx 0.41$. We emphasise that this is a \emph{reference} value rather than a true upper bound: $\gamma_e < 1$ is plausible (the Liu et al.\ width-only estimate gives $\gamma_e \approx 0.40$), and $\beta_s$ is unconstrained. The logistic mapping introduces the additional linearisation factor $a$, which the fitted $\alpha$ also absorbs. With these caveats, three of four within-family log-linear slopes cluster tightly at 0.22--0.24 (Mistral 0.218, Llama 0.224, Gemma 0.238), or 54--59\% of the canonical-reference $\mmax = 0.407$, consistent with models using roughly half their parameter budget for non-factual capabilities relative to that reference. The Qwen3 series is an outlier (slope 0.366, or 90\% of $\mmax$), but this reflects its smallest member (Qwen3-8B, $q = 0.054$) sitting in the floor regime in which the $\Neff \propto P^{\gamma_e}$ assumption breaks down; the steep apparent slope is partly the climb out of that floor rather than a clean within-family scaling estimate (Table~\ref{tab:consistency}). In summary, the Zipf exponent $\aZ$ sets the reference slope against which encoding efficiency is measured.

\begin{table}[ht]
  \centering
  \caption{\textbf{Reference-slope quantities under different Zipf-exponent
  estimates.} Empirical comparators: $m_{\text{Llama}} = 0.224$, $n = 0.080$ (cross-family OLS).
  Efficiency is $m/\mmax$ with $\mmax = 1/(2\aZ)$ at the canonical
  reference $\gamma_e = \beta_s = 1$.}
  \label{tab:consistency}
  \small
  \begin{tabular}{@{}lccc@{}}
    \toprule
    Scenario & $\aZ$ & $\mmax$ & Efficiency \\
    \midrule
    Canonical Zipf        & 1.00 & 0.500 & 44.8\% \\
    \textbf{24-topic OLS} & \textbf{1.23} & \textbf{0.407} & \textbf{55.0\%} \\
    Global MLE            & 1.24 & 0.403 & 55.6\% \\
    \bottomrule
  \end{tabular}
\end{table}

\subsection{Predictions}

\begin{figure}[!ht]
  \centering
  \includegraphics[width=\textwidth]{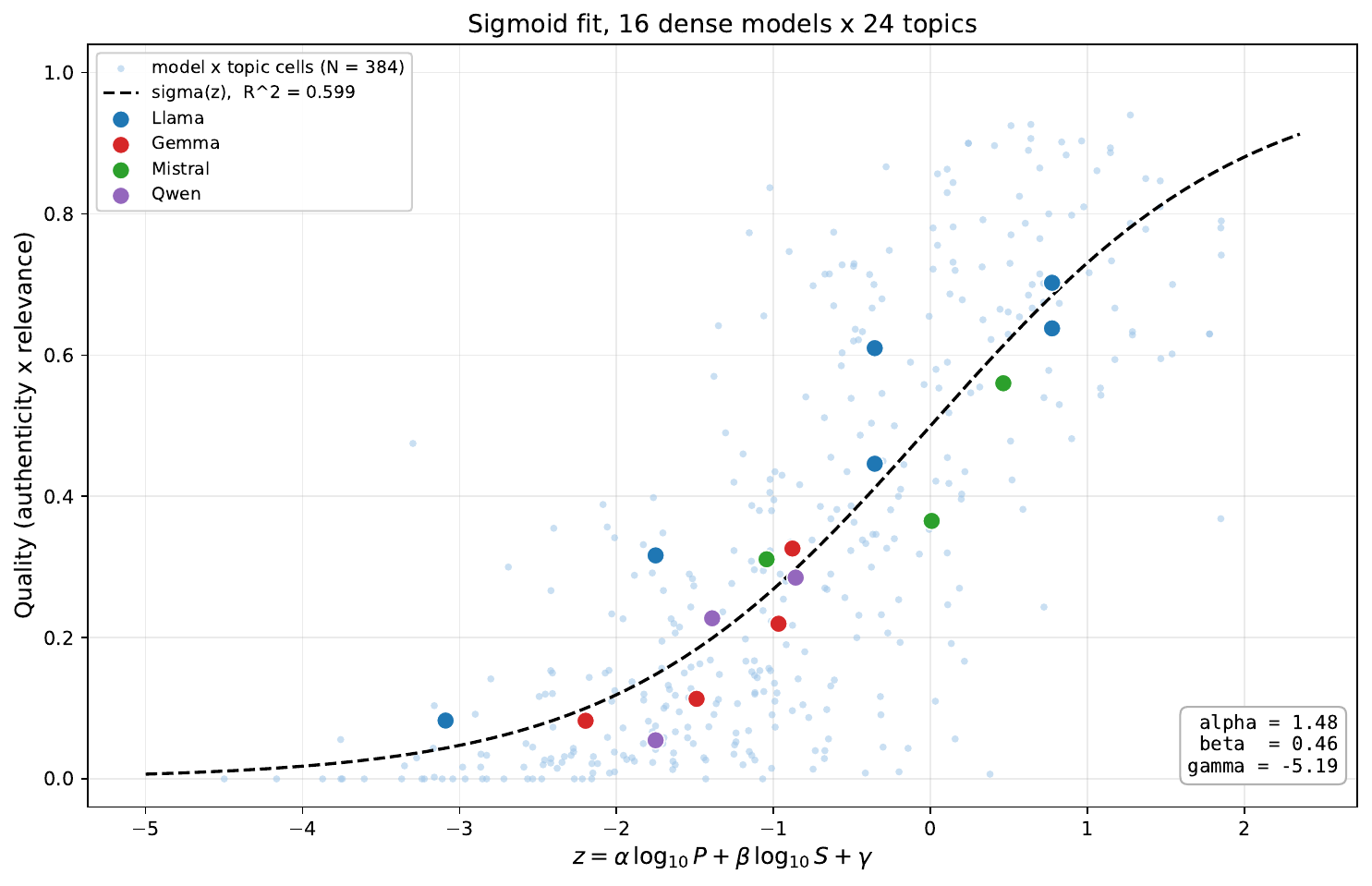}
\caption{\textbf{Factual recall quality follows a sigmoid in the log-linear combination of model size and topic representation frequency.} Light blue points show individual model$\times$topic observations ($N = 384$); coloured points show per-model averages by family (Llama, Gemma, Mistral, Qwen). The dashed curve is the fitted logistic $\sigma(z)$, where $z = \alpha\log_{10}P + \beta\log_{10}S + \gamma$ ($R^2 = 0.599$). Residual variance is greatest in the mid-range where models are near the recall threshold, compressing toward zero at the floor and ceiling as predicted by the sigmoid geometry.}
  \label{fig:sigmoid}
\end{figure}

The framework yields five predictions, each consistent with the data:
\begin{enumerate}[nosep]
  \item Within-family log-linear scaling ($R^{2} = 0.74$--$0.94$ across four families with $\geq 3$ model sizes each: Llama $n = 6$, Gemma $n = 4$, Mistral $n = 3$, Qwen $n = 3$; treated as descriptive rather than inferential given the small per-family $n$).
  \item Cross-topic log-linear scaling: a positive content slope $n$ in Eq.~\ref{eq:quality-model} across all four families (Llama, Gemma, Mistral, Qwen each show $n > 0$).
  \item Gradual degradation across verification categories as capacity decreases.
  \item Empirical slope below the canonical reference: $m < 0.41$.
  \item Sigmoid saturation: quality-vs-$\log P$ traces a logistic curve with floor and ceiling regimes (Fig.~\ref{fig:sigmoid}).
\end{enumerate}


\section{Methods}
\label{sec:methods}

\subsection{Model selection and reference generation}

The primary analysis uses 16 dense models spanning 1B to 405B parameters across four families (Llama, Gemma, Qwen, Mistral), yielding 3,730 evaluated references. An additional 22 model runs (MoE architectures, chain-of-thought variants, and closed-source models with unknown parameter counts) were collected for supplementary analyses, bringing the total to 38 runs and 8,913 evaluated references (Table~\ref{tab:accounting}; Appendix Table~\ref{tab:si-models}).

Each model produced 10 scholarly references for each of 24 topics drawn from six thematic groups, each containing four topics at decreasing specificity (Appendix Table~\ref{tab:si-topics}). This creates a gradient in training-data representation from millions of scholarly works (broad topics) to fewer than 50 (specific topics). All non-thinking models were run at temperature~0 for deterministic output. Thinking models (Qwen3 think, DeepSeek~R1) do not accept a temperature parameter and use the model's default reasoning mode; these variants are excluded from the primary analysis.

The prompt template was identical across all model runs (verbatim, with \texttt{\{topic\}} substituted per request):
\begin{quote}\small\ttfamily\raggedright
List 10 different relevant scholarly references (journal papers, conference papers, technical reports, or dissertations) about \{topic\}\\[2pt]
\#\# RULES:\\
1. Use standard APA citation format: Author(s) (Year). Title. Journal/Publisher.\\
2. Provide 10 distinct references --- no duplicates.\\
3. Only provide the list. No commentary, questions, or explanations.
\end{quote}
The Qwen3 nothink variants additionally prepend the Qwen-specific control token \texttt{/no\_think} on its own line to disable the model's default reasoning mode; the rest of the prompt is unchanged.

\subsection{Reference collection and deduplication}

Reference counts are summarised in Table~\ref{tab:accounting}. Of $9{,}120$ candidate references requested ($38 \times 24 \times 10$), $8{,}913$ were produced and submitted to SourceVerify for authentication (Section~\ref{sec:authentication}); the $207$-reference shortfall reflects cells where the model returned fewer than 10 separable APA citations (concentrated in the smallest models, plus one refusal cell---GPT-5 Nano on ``Biometric voter registration''---and one zero-output cell---DeepSeek V3 on ``School dropout''). A further 84 normalized-title duplicates within model$\times$topic were collapsed during matching, yielding $8{,}829$ analysed references across $910$ of the $912$ populated model--topic cells. Deduplication is performed only within (model, topic) cells: each cell corresponds to a single prompt, so a title repeated within one model's response to one topic is one recall event scored once, whereas the same title produced under a different topic prompt is treated as a separate recall event because it responds to a different question. Title normalization for the dedup match strips case, punctuation, and whitespace, so minor formatting drift is collapsed. For the 16 dense models used in the primary sigmoid fit, $3{,}730$ of $3{,}840$ candidates (97.1\%) were produced and all 16 models covered all 24 topics, so the $N = 384$ model--topic cell structure is preserved.

\begin{table}[ht]
  \centering
  \caption{\textbf{Reference counts from request to analysis.} Counts are exact and reconcile in both columns: full set $9{,}120 - 207 - 84 = 8{,}829$; dense-16 subset $3{,}840 - 110 - 69 = 3{,}661$. The dense-16 subset is the basis for the primary sigmoid fit (Fig.~\ref{fig:sigmoid}).}
  \label{tab:accounting}
  \small
  \begin{tabular}{@{}lrrp{0.40\linewidth}@{}}
    \toprule
    \textbf{Stage} & \textbf{Full set (38)} & \textbf{Dense-16 subset} & \textbf{Loss source} \\
    \midrule
    Requested              & 9{,}120 & 3{,}840 & 38 (16) models $\times$ 24 topics $\times$ 10 references \\
    Produced and evaluated & 8{,}913 ($-207$) & 3{,}730 ($-110$) & Cells returning $<10$ separable APA citations; 1 refusal cell (GPT-5 Nano, biometric voter registration); 1 zero-output cell (DeepSeek V3, school dropout) \\
    Analysed               & 8{,}829 ($-84$)  & 3{,}661 ($-69$)  & Normalized-title duplicates collapsed within model$\times$topic \\
    \bottomrule
  \end{tabular}
\end{table}

\subsection{Authentication}
\label{sec:authentication}

Each reference was submitted to SourceVerify\cite{sourceverify}, which (i) extracts the key reference details, (ii) searches OpenAlex, Google Scholar, Google Books, and DOI registries for candidate matches, and (iii) returns field-level comparisons for title, authors, year, venue, and identifier. Each field is classified as \textsc{match}, \textsc{abbrev}, \textsc{contains}, \textsc{contradiction}, \textsc{unconfirmed}, or \textsc{absent}. We compute a continuous authenticity score as
\begin{equation}
    \text{authenticity} = \max\!\left(0,\;
      \frac{\sum_i w_i \cdot v_i}{\sum_i w_i}
    \right) \quad \text{(non-\textsc{absent} fields)}
    \label{eq:score}
\end{equation}
with field weights title (0.25), identifier (0.25), authors (0.20), year (0.15), venue (0.15) and verdict scores \textsc{match} $= +1.0$, \textsc{abbrev} $= +0.75$, \textsc{contains} $= +0.5$, \textsc{unconfirmed} $= 0.0$, \textsc{contradiction} $= -1.0$. \textsc{Absent} fields are excluded from both numerator and denominator. Thus, a contradicted field counts negatively, whereas a field SourceVerify could not assess does not penalize the score.

\subsection{Authentication validation}

\begin{figure}[h]
  \centering
  \includegraphics[width=0.7\linewidth]{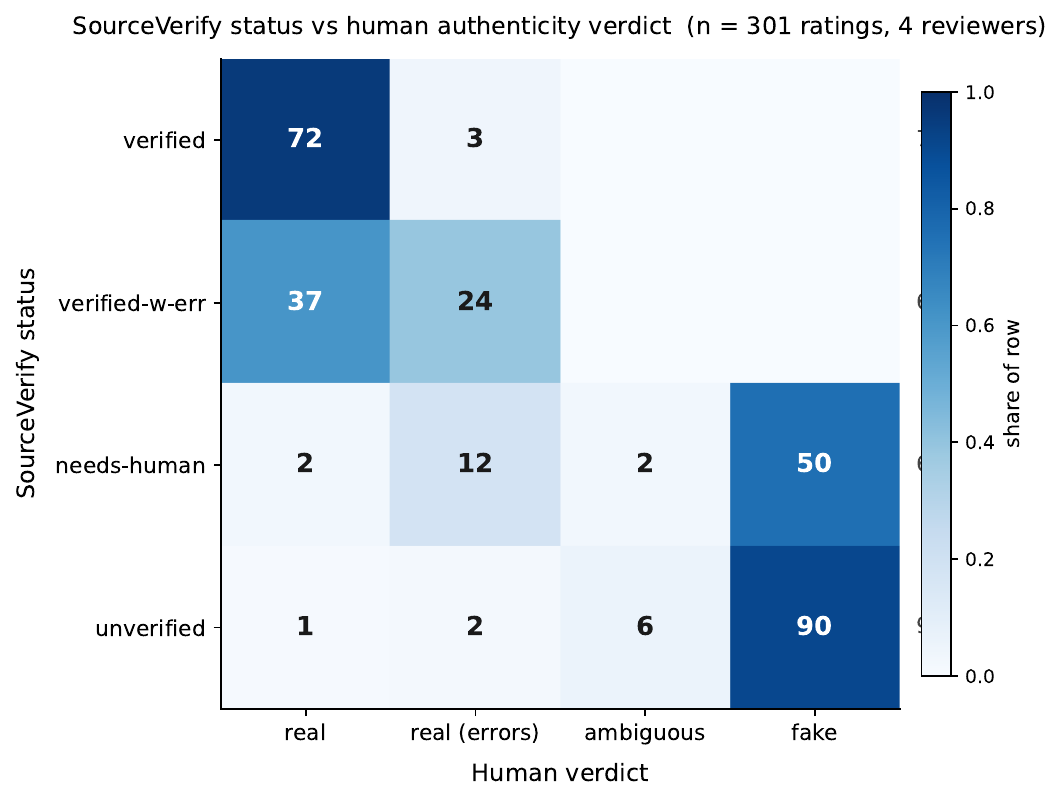}
  \caption{\textbf{SourceVerify status versus human authenticity verdict.}
  Confusion matrix for 301 independent ratings (288 unique references; 13 double-rated) by four human reviewers across four SourceVerify status categories. Both \textit{verified} (75/75) and \textit{verified-with-error} (61/61) contain exclusively real papers (100\%); \textit{unverified} is 97\% not real with 3 real papers missed; \textit{needs-human} is the genuine grey zone (52/66 not real). Treating \textit{ambiguous} human verdicts as not real ($n = 301$), binary precision is 100\% and specificity is 100\%; all 17 disagreements are SourceVerify false negatives.} \label{fig:confusion}
\end{figure}

A total of 288 references were rated by four human reviewers (301 ratings; 13 double-rated, with 100\% pairwise binary agreement). Human verdicts were \textit{real} (yes or yes-with-errors), \textit{ambiguous}, or \textit{fake} (no); treating \textit{ambiguous} as fake, SourceVerify agreed with humans on 94.4\% of ratings (Cohen's $\kappa = 0.887$; recall $88.9\%$, precision and specificity both $100\%$; Fig.~\ref{fig:confusion}).

The four SourceVerify status categories separate cleanly by authenticity (Fig.~\ref{fig:confusion}): \textit{verified} and \textit{verified-with-error} contain only real papers, \textit{unverified} contains almost exclusively non-real papers, and \textit{needs-human} is the genuine grey zone where human adjudication adds value. All 17 human--SourceVerify disagreements were false negatives (real papers SourceVerify could not confirm). All 17 of these references also had at least one factual defect (wrong venue, year, coauthor, or drifted title). SourceVerify never certified a fabricated reference as real, so reported authenticity and quality scores cannot be inflated by retrieval false positives.

\subsection{Relevance scoring}

Each reference was judged for topic relevance by Kimi~K2 (32B active MoE, temperature~0) on a three-level scale: YES (1.0), PARTIAL (0.50), or NO (0.0).

Two independent human raters ($n = 347$ ratings across 295 stratified references) achieved 96.3\% binary agreement with the automated judge (Cohen's $\kappa = 0.394$ three-level, weighted $\kappa = 0.432$). Nearly all disagreement occurred on the YES-vs-PARTIAL boundary; K2 never rated YES when a human rated NO (0/347).

A sweep of PARTIAL weights from 0.0 to 1.0 confirmed robustness to this choice: the model rank order against the PARTIAL$=0.50$ baseline is preserved (Spearman $\rho \geq 0.997$ across the full range; Appendix Table~\ref{tab:si-robustness}), with sigmoid $R^{2}$ ranging from 0.504 to 0.606 around the main-fit value of 0.599 (log-linear $R^{2}$ from 0.481 to 0.573). Combined quality is the product of authenticity and relevance: $\text{quality} = \text{authenticity} \times \text{relevance}$.

\subsection{Statistical methods}

OpenAlex scholarly work counts per topic (queried 8 March 2026 via the \texttt{title\_and\_abstract.search} filter) served as the proxy for training-data representation ($S$ in Equation~\ref{eq:model}). Log-linear fits use ordinary least squares in $\log_{10}$ space; the sigmoid fit uses nonlinear least squares (Levenberg--Marquardt). Within-family fits used $n = 6$ model sizes (Llama, including two 70B variants and the 405B base together with its Hermes fine-tune), $n = 4$ (Gemma), or $n = 3$ (Mistral, Qwen). Cross-family fits included only dense models to avoid confounding by MoE routing variance. Reported standard errors and $R^2$ values from the sigmoid fit assume independent model--topic residuals; clustering by model and by topic group would yield wider intervals, and we accordingly treat the reported uncertainties as point-estimate descriptive summaries rather than inferential bounds.

The continuous authenticity score and a binary title-match alternative (a single yes-or-no question requiring no field-level judgements) yield near-identical sigmoid fits ($R^{2} = 0.599$ and $0.604$ respectively), suggesting that the moderate explained variance reflects the underlying phenomenon rather than measurement choice. The remaining ${\sim}40\%$ of unexplained variance is attributable to factors beyond parameter count and topic frequency: models with the same parameter count but different training (e.g.\ Llama~405B base vs Hermes fine-tune, or Llama~8B vs Qwen3~8B nothink) show quality differences of up to 0.23 that neither $P$ nor $S$ can capture.

\subsection{Citation-count analysis}

For correctly recalled references, the number of times each has been cited (its citation count) provides a within-topic test of the training-data-representation hypothesis: if recall is driven by frequency in training data, then as model capacity grows, the marginal recalled reference should sit further down the citation tail. Higher-cited (better-represented) papers are recalled first, lower-cited (less-represented) ones only once the noise floor drops. Inclusion was restricted to references with SourceVerify status \textit{verified} or \textit{verified-with-error} (per the IRR audit, both buckets are 100\% real papers). For each such reference, we extracted the paper title from the model's APA-formatted output and queried the OpenAlex \texttt{/works} endpoint, accepting the top-ranked result only when content-word overlap (a small stopword list stripped before comparison) was at least 50\%. The reported citation count is OpenAlex's \texttt{cited\_by\_count} field; OpenAlex undercounts books and grey literature relative to Google Scholar, but the relative ranking of works by citation impact is reliable. Two exclusions apply: $\sim 10\%$ of candidate references could not be confidently matched in OpenAlex (no hit, or low-overlap top hit), which biases reported medians upward (unmatched papers are disproportionately obscure) and makes the small-vs.-large-model contrast a conservative lower bound; and models contributing fewer than 50 matched references were excluded to keep the per-model median estimable. Per-model 95\% confidence intervals on the median were obtained by $10{,}000$-resample bootstrap, and the cross-model log--log fit weighted each point by the inverse variance of its bootstrap-derived standard error.


\section{Results}
\label{sec:results}

\subsection{Log-linear scaling with model size}

\begin{figure}[!ht]
  \centering
  \includegraphics[width=\linewidth]{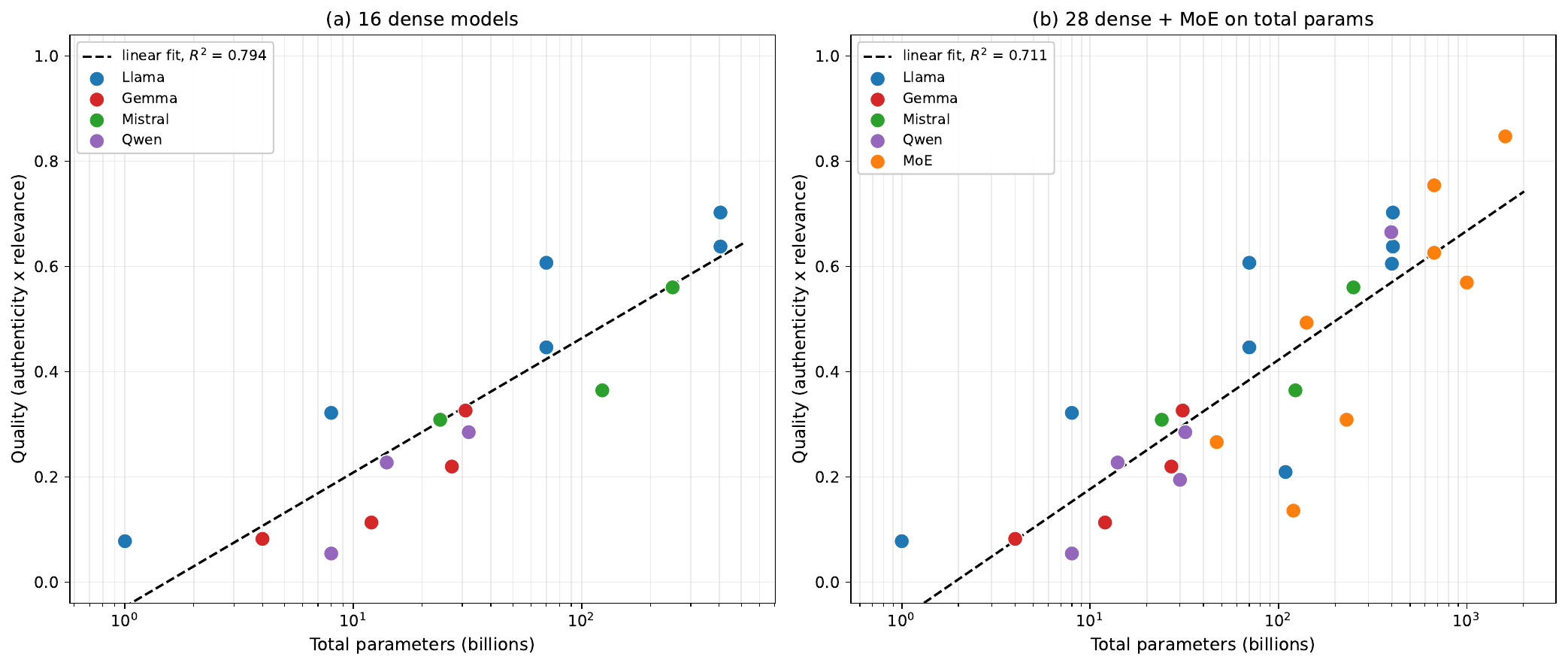}
 \caption{\textbf{Factual recall quality scales log-linearly with model size across architectures and training generations (e.g., Llama 3.1, 3.2, \& 3.3).} (a)~Quality versus parameter count for \textbf{16} dense models from four independent families (Llama, Gemma, Mistral, Qwen). Dashed line shows log-linear fit ($R^2 = 0.794$). Systematic offsets between families reflect differences in training procedure rather than scale. (b)~The relationship extends to \textbf{28} models across all architectures when plotted against total parameter count ($R^2 = 0.712$). MoE models (diamonds) follow the same trend as dense models, suggesting total rather than active parameters dominate the noise floor.}
  \label{fig:scaling}
\end{figure}

Factual recall quality increases log-linearly with parameter count across 16 dense models spanning 1B to 405B ($R^2 = 0.794$; Fig.~\ref{fig:scaling}a). The relationship holds across four independent model families with clear systematic offsets. Llama models consistently exceed the cross-family trend while Gemma and Qwen models fall below it, indicating that factors beyond parameter count, such as training procedure or data curation, shift the intercept without changing the slope. Extending to 28 models across dense and Mixture of Expert (MoE) architectures, with MoE models plotted on total rather than active parameter count, preserves the log-linear trend ($R^2 = 0.712$; Fig.~\ref{fig:scaling}b), consistent with total parameters, rather than active parameters, dominating the interference noise floor under a superposition account.

\subsection{Topic representation and recall quality}

\begin{figure}[t]
  \centering
  \includegraphics[width=\columnwidth]{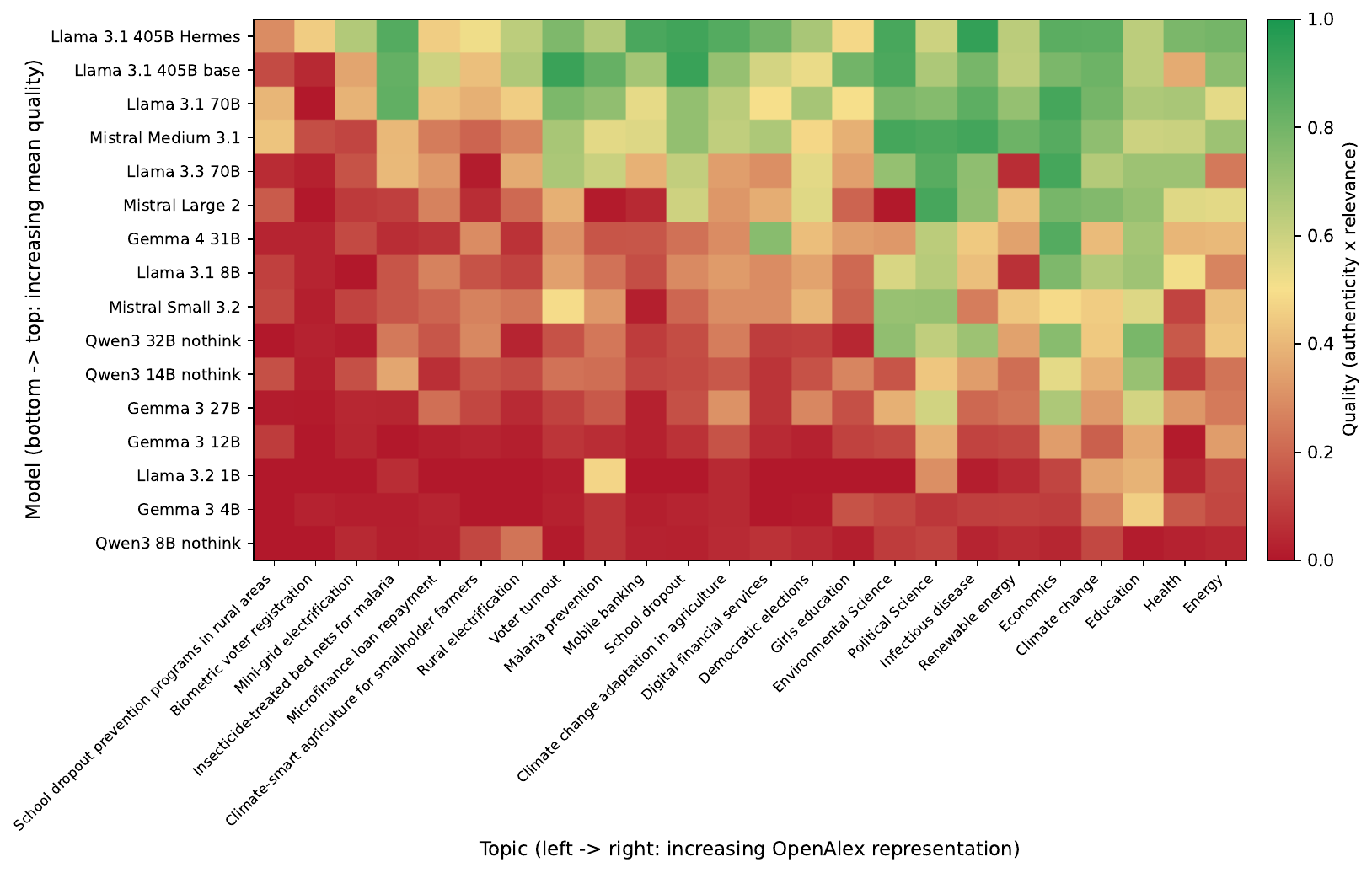}
  \caption{\textbf{Topic representation drives recall quality across model families.} Recall quality across 16 dense models and 24 topics, sorted by topic representation frequency (left to right) and model quality (bottom to top). Colour encodes quality (0--1), with red indicating low recall, yellow the 0.5 threshold, and green high recall. The gradient is consistent with topic frequency being positively associated with recall quality across model families.}
  \label{fig:topicfreq}
\end{figure}

Topic representation is associated with recall quality across nearly all 38 models tested, independent of architecture or parameter count. 37 of 38 models show a positive Spearman correlation between quality and $\log_{10}S$ (median $\rho = 0.55$); the single outlier is Grok~3 ($\rho = -0.04$). The near-zero correlation for DeepSeek V4 Pro ($\rho = 0.02$) is consistent with ceiling compression: at quality $= 0.847$, nearly all references are correct regardless of topic frequency, attenuating the correlation (Fig.~\ref{fig:topicfreq}).

\subsection{Sigmoid fit}

We fit a sigmoid to $N = 384$ model--topic observations from 3{,}661 references generated by 16 dense models spanning 1B to 405B parameters across 24 topics. $\alpha = 1.48 \pm 0.09$, $\beta = 0.46 \pm 0.04$, $\gamma = -5.19 \pm 0.31$ (all $|t| > 11$, $p \approx 0$); $R^2 = 0.599$. Model size alone ($\log_{10} P$) accounts for 42.1\% of variance; topic frequency ($\log_{10} S$) adds a further 17.8\% ($F_{1,381} = 169.6$, $p \approx 0$), so both axes contribute independently to recall quality. We caution that the standard errors and $F$-statistic above treat the 384 model--topic observations as independent; clustering by model and by topic group would yield wider intervals (see Section~\ref{sec:methods}).

\subsection{Robustness: binary title-match metric}

To test whether the unexplained variance lies in the measurement rather than the phenomenon, we simplified the scoring to a single binary question: \textit{does the model recall the exact title of a real paper}? Fitting the same sigmoid to this binary title-match metric yields $\alpha = 2.08 \pm 0.15$, $\beta = 0.52 \pm 0.06$, $\gamma = -6.05 \pm 0.44$ (all $|t| > 9$) with $R^2 = 0.604$, similar to the continuous score despite using strictly less information. The near-identical fit suggests that the moderate explained variance reflects the underlying phenomenon rather than measurement noise. The marginally steeper $\alpha$ (2.08 vs 1.48) is consistent with binary title recall being more sensitive to model size than the continuous composite score, as expected for an all-or-nothing retrieval metric.

\subsection{Citation-count gradient}

\begin{figure}[t]
\centering
\includegraphics[width=0.65\linewidth]{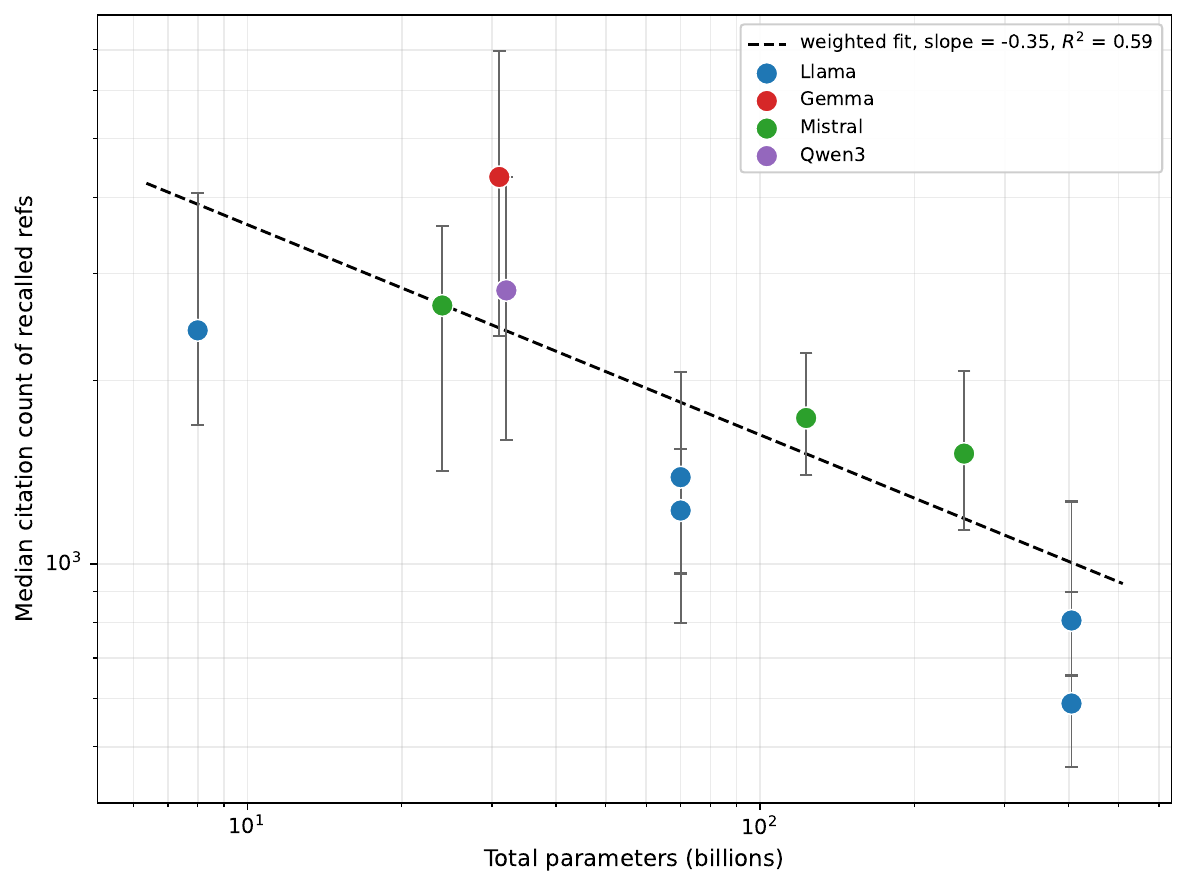}
\caption{\textbf{Larger models recall less-cited papers.} Median citation count of correctly recalled references (SourceVerify status \textit{verified} or \textit{verified-with-error}) plotted against model size on a log--log scale, for 10 dense models from four families with $n \geq 50$ matched references each. Error bars are bootstrap 95\% CIs on the per-model median ($10{,}000$ resamples). The dashed line is a weighted log--log fit using the bootstrap-derived standard errors as inverse-variance weights (slope $= -0.35$, $R^2 = 0.59$, Spearman $\rho = -0.79$, $p = 0.007$). Larger models extend recall further down the citation tail, consistent with a superposition account in which stronger parameter redundancy is required to encode low-frequency information.}
\label{fig:citations}
\end{figure}

If highly cited works are referenced more often in training data, their bibliographic details should be encoded with stronger parametric redundancy. Recall gated by signal strength then predicts that small models will recall only the most-cited papers while larger models extend recall down the citation tail. Among correctly recalled references (SourceVerify status \textit{verified} or \textit{verified-with-error}) that matched a unique OpenAlex record, the median citation count decreases log-linearly with model size across 10 dense models from four families with $n \geq 50$ matches each (weighted log--log slope $= -0.35$, $R^2 = 0.59$, Spearman $\rho = -0.79$, $p = 0.007$). Llama~3.1~8B recalls papers with median 2{,}419 citations, while Llama~3.1~405B drops to 806 (base) or 589 (Hermes fine-tune), with bootstrap 95\% CIs on every per-model median (Fig.~\ref{fig:citations}).

\subsection{Three regimes of the sigmoid}

The sigmoid fit consists of three qualitatively distinct regimes. In the \textit{floor} regime, models produce templated fabrications. In the \textit{ramp} regime, recall quality increases log-linearly with both model size and topic representation. In the \textit{ceiling} regime, gains saturate as capacity exceeds the concept inventory.

At the bottom of the sigmoid, models stop producing references and begin producing \textit{templates}. Llama~3.2~1B produced 215 references, of which only 22 were verifiable; eight of its nine Political Science references are the same work, Dahl's \textit{Polyarchy} (1971), regenerated with different publication years. Qwen3~8B draws from a pool of only 62 unique first-author surnames across 236 references, with ``Smith'' appearing in all 24 topics. These are not necessarily researchers being recalled, but high-frequency surnames in the training corpus recycled as slot-fillers.

The transition across regimes is gradual rather than binary: Llama~405B (base) produces 72\% fully verified references and 13\% unverified; at 70B these shift to 63\% and 18\%; at 8B, only 30\% are verified while 42\% are entirely unverified. Intermediate categories persist across all sizes, indicating a regime where models retrieve real papers but corrupt specific fields. This mid-range is also genuinely ambiguous because the continuous score conflates corrupted recall of a real source with coincidental overlap with an unrelated one, introducing measurement noise precisely where the sigmoid is steepest. Residual variance is greatest in the middle prediction third (SD~$= 0.181$, $1.7\times$ the low band at 0.108) and the high band (0.200), consistent with the steepest part of the sigmoid having the most measurement-induced noise and ceiling-side compression flattening the high band.

At the top of the sigmoid, the largest closed-source models (Claude Opus~4.6, GPT-5, GPT-5.4) plateau at 0.76--0.81 quality, while DeepSeek V4 Pro, an open-weights MoE model with 1.6T total parameters, exceeds this range at 0.847. This suggests that the ceiling regime may not yet have been reached for the largest open-weights models. Closed-source comparisons are necessarily exploratory because exact parameter counts are undisclosed.


\section{Discussion}
\label{sec:discussion}

These results extend the scaling-law framework of \citet{kaplan2020} and \citet{hoffmann2022} from aggregate performance to factual recall, providing a unified functional form and quantifying the role of topic frequency as a second independent predictor. That aggregate gains reflect improved factual encoding is expected under superposition theory\cite{elhage2022, liu2025superposition}, since larger models sustain lower interference floors, bringing progressively more concepts into reliable recall. What was not previously established is that, across four independent model families, the resulting recall quality follows a sigmoid in the latent log signal-to-noise ratio. This treats factual confabulation as a tractable consequence of compressing non-uniformly distributed knowledge into finite representational capacity.

As model capacity decreases, the noise floor rises and concept representations are progressively overwhelmed, with rarer concepts falling below the recall threshold first. This account also explains the systematic intercept offsets between model families. Models trained on cleaner or more domain-relevant data produce stronger parametric representations, shifting the sigmoid leftward without changing its slope. Confabulations are therefore not random but reflect a structural inequality in how knowledge is encoded, consistent with the formal result that hallucination is an inherent limitation of LLMs\cite{xu2024} and with recent evidence of systematic geographic disparities in LLM factual recall\cite{moayeri2024worldbench}: high-frequency concepts sit above the interference threshold and are consistently retrieved, while low-frequency concepts remain near or below it.

Scaling lowers the noise floor, bringing progressively rarer concepts into reliable recall, but this is coverage expansion, not uniform improvement. Gains for already well-represented topics saturate, while extending reliable recall into the long tail requires disproportionate increases in capacity. The topics in this study span more than five orders of magnitude in training-data representation, from \textit{climate change} ($S \approx 1{,}200{,}000$) to \textit{school dropout prevention in rural areas} ($S = 32$). DeepSeek V4 Pro, the largest model tested at 1.6T total parameters, achieves a quality score of 0.90 on the former but only 0.43 on the latter. Solving Equation~\ref{eq:model} for $P$ at fixed quality suggests that, under the fitted dense-model relationship, raising the score on the low-frequency topic to 0.90 would require approximately 50T parameters, roughly 30$\times$ the scale of the largest model tested, and likely more in practice, since MoE architectures in our experiments underperform the dense sigmoid at equivalent total parameter counts. Aggregate benchmarks that average across the frequency spectrum therefore overestimate reliability for long-tail concepts and mask systematic disparities across topics, geographies, and languages.

Data curation, targeted pre-training, and inference-time interventions should show the largest gains for concepts near the noise floor, where a modest increase in signal suffices to cross the recall threshold. For very low-frequency concepts below the floor, retrieval augmentation that bypasses parametric recall entirely is the appropriate response. A direct test of the causal role of training data frequency would require controlled pre-training experiments in which concept representation is varied independently of model capacity, something the present observational design cannot support. The sigmoid parameterisation nonetheless offers a quantitative way to estimate the increase in $\log_{10} S$ required to push a concept across the recall threshold ($\sigma(z) = 0.5$) at a given model size, computed as $\Delta\log_{10}S = -(\hat{\alpha}\log_{10}P + \hat{\gamma})/\hat{\beta}$, from the fitted parameters.


\section{Limitations}
\label{sec:limitations}

\paragraph{Single factual domain.}
Scholarly references are well-structured, externally indexed, and amenable to automated verification. Whether the sigmoid scaling relationship generalises to other forms of factual knowledge---commonsense facts, numerical quantities, procedural knowledge---remains untested. References may also be overrepresented in LLM training corpora relative to other factual domains, so the fitted parameters ($\alpha$, $\beta$, $\gamma$) should not be taken as universal constants.

\paragraph{Training-data opacity.}
The topic representation proxy $S$ (OpenAlex work counts) is an indirect measure of training-data frequency. We do not have access to the training corpora of any model tested, so the assumption $f \propto S^\delta$ is empirically motivated but not directly verified. Models with different data-curation pipelines may induce different effective $\delta$ values, contributing to the observed family-level intercept offsets.

\paragraph{Observational design.}
The study is correlational: model size and training-data quality co-vary across model generations, and we cannot manipulate concept frequency independently of model capacity. Establishing a causal role for training-data representation would require controlled pre-training experiments.

\paragraph{Within-family sample sizes.}
The primary sigmoid fit pools 16 models, but within-family fits rely on as few as $n = 3$ model sizes (Mistral, Qwen). These slopes are therefore indicative rather than precise, and the Qwen3 outlier slope illustrates how a single model in the floor regime can dominate a small-sample fit.

\paragraph{English only.}
All topics and references are in English. For low-resource languages, training-data representation is orders of magnitude lower, and the scaling relationship could shift substantially in both slope and intercept.

\paragraph{Fixed prompt and decoding.}
All non-thinking models were evaluated with a single prompt template at temperature~0. Different prompting strategies (few-shot, chain-of-thought) or non-zero temperature could shift the sigmoid parameters, particularly for models near the recall threshold.


\section{Conclusion}
\label{sec:conclusion}

We have shown that factual recall quality in large language models follows a sigmoid in the log-linear combination of model parameter count and topic representation in training data. The functional form, motivated by a superposition-inspired signal-to-noise account, fits data from 16 dense models across four independent families ($R^2 = 0.599$) and generalises across 38 models spanning dense, MoE, and closed-source architectures. The framework reveals three regimes---a floor of templated fabrication, a ramp of log-linear improvement, and a ceiling of saturation---and predicts that larger models extend recall into the citation tail rather than uniformly improving across all topics.

These findings reframe factual confabulations as a predictable consequence of compressing non-uniformly distributed knowledge into finite representational capacity, rather than as random or idiosyncratic failures. Natural extensions include other factual domains beyond scholarly references, multilingual settings, and the interaction between parametric recall and retrieval augmentation.


\section*{Acknowledgements}

This work was carried out with support from the Artificial Intelligence for Development (AI4D) Africa programme, with financial support from Canada's International Development Research Centre (IDRC) and the UK's Foreign, Commonwealth \& Development Office (FCDO). The views expressed herein do not necessarily represent those of IDRC or its Board of Governors, or those of FCDO.


\paragraph{Data availability.}
Verified reference data for all open-source model runs, relevance judgements, and topic metadata are available at \url{https://github.com/matthewlongshore/predictable-confabulations}. Closed-source model outputs (GPT-5 family, Claude family) are included as aggregate scores but raw outputs are not re-distributable under API terms.

\paragraph{Code availability.}
All analysis scripts for verification, scoring, sigmoid fitting, and figure generation are available at the same repository.

\section*{Author Contributions}
M.S.\ designed the study, collected model outputs, ran verifications, performed the primary data analysis, and helped to write the manuscript. J.S.\ developed the theoretical framework, performed the sigmoid and scaling analysis, and wrote the manuscript. S.T.S., T.F.B., and I.E.O.\ contributed to study design and manuscript revision. All authors reviewed and approved the final manuscript.

Use of AI tools. During preparation of this manuscript, the authors used Anthropic's Claude (Opus and Sonnet models, accessed November 2025--May 2026) to assist with manuscript editing, consistency checks across drafts, simulated peer review of theoretical claims, and LaTeX debugging. The model was also used to draft analysis scripts that the authors reviewed, executed, and verified independently. All scientific claims, the experimental design, the data collection and verification pipeline, and the theoretical framework were developed by the authors. The authors reviewed and edited all AI-assisted text and take full responsibility for the content of the manuscript.

\section*{Competing Interests}
M.S.\ is a co-founder of SourceVerify, the reference verification service used to grade the model outputs in this study, and therefore has a financial interest in its success. To mitigate this conflict, the inter-rater reliability evaluation (Methods, ``Authentication validation'') was performed independently by the other four authors (J.S., S.T.S., T.F.B., and I.E.O.); M.S.'s ratings were excluded from all reported precision, recall, specificity, accuracy, and Cohen's $\kappa$ figures. The downstream sigmoid fitting and scaling analyses were conducted on the verified reference outputs without further input from SourceVerify, and the full reference-level dataset (including raw verdict fields and human-review labels) is released alongside the paper to permit independent re-analysis. The remaining authors declare no competing interests.



\appendix
\renewcommand{\thetable}{A\arabic{table}}
\renewcommand{\thefigure}{A\arabic{figure}}

\section{Supplementary Tables}

\begin{table}[H]
  \centering
  \tiny
  \caption{\textbf{All 38 models tested.} Active parameters listed for MoE; total in parentheses. Quality = authenticity $\times$ relevance. Scores use contradiction penalty ($-1.0$).}
  \label{tab:si-models}
  \renewcommand{\arraystretch}{1.10}
  \begin{tabular}{llrcrl}
    \toprule
    \textbf{Model} & \textbf{Family} & \textbf{Params} & \textbf{Auth.}
      & \textbf{Quality} & \textbf{Arch.} \\
    \midrule
    Llama 3.2 1B          & Llama    & 1B          & 0.087 & 0.078 & Dense \\
    Gemma 3 4B            & Gemma    & 4B          & 0.100 & 0.082 & Dense \\
    Llama 3.1 8B          & Llama    & 8B          & 0.425 & 0.322 & Dense \\
    Qwen3 8B (nothink)    & Qwen     & 8B          & 0.067 & 0.054 & Dense \\
    Qwen3 8B (think)      & Qwen     & 8B          & 0.139 & 0.107 & Dense+CoT \\
    Gemma 3 12B           & Gemma    & 12B         & 0.143 & 0.113 & Dense \\
    Qwen3 14B (nothink)   & Qwen     & 14B         & 0.274 & 0.227 & Dense \\
    Qwen3 14B (think)     & Qwen     & 14B         & 0.219 & 0.174 & Dense+CoT \\
    Mistral Small 3.2     & Mistral  & 24B         & 0.396 & 0.309 & Dense \\
    Gemma 3 27B           & Gemma    & 27B         & 0.273 & 0.220 & Dense \\
    Gemma 4 31B           & Gemma    & 31B         & 0.365 & 0.326 & Dense \\
    Qwen3 32B (nothink)   & Qwen     & 32B         & 0.362 & 0.285 & Dense \\
    Qwen3 32B (think)     & Qwen     & 32B         & 0.263 & 0.213 & Dense+CoT \\
    Llama 3.1 70B         & Llama    & 70B         & 0.724 & 0.607 & Dense \\
    Llama 3.3 70B         & Llama    & 70B         & 0.570 & 0.446 & Dense \\
    Mistral Large 2       & Mistral  & 123B        & 0.449 & 0.364 & Dense \\
    Mistral Medium 3.1    & Mistral  & 250B        & 0.678 & 0.560 & Dense \\
    Llama 3.1 405B (base) & Llama    & 405B        & 0.779 & 0.638 & Dense \\
    Llama 3.1 405B Hermes & Llama    & 405B        & 0.879 & 0.703 & Dense (ft) \\
    \midrule
    GPT-OSS 120B          & ---      & 5.1B (120B) & 0.162 & 0.136 & MoE \\
    Qwen3 30B-A3B         & Qwen     & 3B (30B)    & 0.215 & 0.195 & MoE \\
    MiniMax M2.5          & ---      & 10B (230B)  & 0.354 & 0.309 & MoE \\
    Mixtral 8x7B          & Mistral  & 13B (47B)   & 0.304 & 0.266 & MoE \\
    Llama 4 Scout         & Llama~4  & 17B (109B)  & 0.269 & 0.209 & MoE \\
    Llama 4 Maverick      & Llama~4  & 17B (400B)  & 0.735 & 0.605 & MoE \\
    Qwen3.5               & Qwen     & 17B (397B)  & 0.711 & 0.665 & MoE \\
    Kimi K2               & ---      & 32B (1T)    & 0.672 & 0.570 & MoE \\
    Mixtral 8x22B         & Mistral  & 39B (141B)  & 0.571 & 0.493 & MoE \\
    DeepSeek R1           & DeepSeek & 37B (671B)  & 0.758 & 0.626 & MoE+CoT \\
    DeepSeek V3           & DeepSeek & 37B (671B)  & 0.851 & 0.754 & MoE \\
    DeepSeek V4 Pro       & DeepSeek & 49B (1600B) & 0.891 & 0.847 & MoE \\
    \midrule
    GPT-5 Nano            & GPT      & unknown     & 0.548 & 0.432 & unknown \\
    GPT-5 Mini            & GPT      & unknown     & 0.795 & 0.623 & unknown \\
    Grok 3                & ---      & unknown     & 0.749 & 0.690 & unknown \\
    GPT-5                 & GPT      & unknown     & 0.876 & 0.762 & unknown \\
    GPT-5.4               & GPT      & unknown     & 0.884 & 0.761 & unknown \\
    Claude Sonnet 4.6     & Claude   & unknown     & 0.861 & 0.693 & unknown \\
    Claude Opus 4.6       & Claude   & unknown     & 0.924 & 0.806 & unknown \\
    \bottomrule
  \end{tabular}
\end{table}

\begin{table}[H]
  \centering
  \caption{\textbf{The 24 research topics} organised by thematic group and specificity level. OpenAlex work counts (queried 8 March 2026) serve as proxy for training-data representation.}
  \label{tab:si-topics}
  \small
  \renewcommand{\arraystretch}{1.10}
  \begin{tabular}{llr}
    \toprule
    \textbf{Group} & \textbf{Topic} & \textbf{OpenAlex Works} \\
    \midrule
    Economics        & Economics                                    & 767{,}282 \\
                     & Digital financial services                   & 34{,}424 \\
                     & Mobile banking                               & 17{,}971 \\
                     & Microfinance loan repayment                  & 1{,}341 \\
    \midrule
    Education        & Education                                    & 5{,}921{,}253 \\
                     & Girls education                              & 71{,}707 \\
                     & School dropout                               & 19{,}413 \\
                     & School dropout prevention in rural areas     & 32 \\
    \midrule
    Energy           & Energy                                       & 8{,}631{,}334 \\
                     & Renewable energy                             & 501{,}950 \\
                     & Rural electrification                        & 9{,}391 \\
                     & Mini-grid electrification                    & 747 \\
    \midrule
    Environ.\ Sci.   & Climate change                               & 1{,}222{,}665 \\
                     & Environmental science                        & 247{,}835 \\
                     & CC adaptation in agriculture                 & 25{,}815 \\
                     & Climate-smart agriculture for smallholders   & 1{,}406 \\
    \midrule
    Health           & Health                                       & 8{,}504{,}910 \\
                     & Infectious disease                           & 469{,}401 \\
                     & Malaria prevention                           & 13{,}463 \\
                     & Insecticide-treated bed nets for malaria     & 1{,}331 \\
    \midrule
    Political Sci.   & Political science                            & 285{,}942 \\
                     & Democratic elections                         & 48{,}258 \\
                     & Voter turnout                                & 10{,}283 \\
                     & Biometric voter registration                 & 171 \\
    \bottomrule
  \end{tabular}
\end{table}

\begin{table}[H]
  \centering
  \caption{\textbf{Geometric interpretation of the scaling law.}}
  \label{tab:si-geometric}
  \small
  \begin{tabular}{lll}
    \toprule
    \textbf{Term} & \textbf{Interpretation} & \textbf{Determined by} \\
    \midrule
    $P$ & Noise floor $1/\!\sqrt{N}$ & Architecture \\
    $S$ & Signal amplitude & Training data \\
    $m$ & Encoding efficiency & Training procedure \\
    $n$ & Signal sensitivity & $\approx$Constant (dense) \\
    $c$ & Baseline SNR offset & Data quality + procedure \\
    \bottomrule
  \end{tabular}
\end{table}

\begin{table}[H]
  \centering
  \caption{\textbf{Relevance threshold robustness.} Key metrics across PARTIAL
  weight values ($N = 384$ model--topic observations, 16 dense models with
  known parameter counts). Spearman $\rho$ is the rank correlation of model
  orderings against the PARTIAL = 0.50 baseline.}
  \label{tab:si-robustness}
  \small
  \begin{tabular}{lccc}
    \toprule
    \textbf{PARTIAL wt} & \textbf{sigmoid $R^{2}$} & \textbf{log-linear $R^{2}$} & \textbf{Spearman $\rho$} \\
    \midrule
    0.00 & 0.504 & 0.481 & 1.000 \\
    0.25 & 0.565 & 0.536 & 1.000 \\
    0.50 & 0.599 & 0.566 & 1.000 \\
    0.75 & 0.606 & 0.573 & 0.997 \\
    1.00 & 0.594 & 0.564 & 0.997 \\
    \bottomrule
  \end{tabular}
\end{table}


\bibliographystyle{unsrtnat}
\bibliography{references}

\end{document}